# Reasoning about RoboCup Soccer Narratives

**Hanneh Hajishirzi**[1], **Julia Hockenmaier**[1], **Erik T. Mueller**[2], **and Eyal Amir**[1]
[1]{hajishir, eyal, juliahmr}@illinois.edu, [2]etm@us.ibm.com
[1]University of Illinois at Urbana-Champaign, [2]IBM Watson Research Center

## Abstract

This paper presents an approach for learning to translate simple narratives, i.e., texts (sequences of sentences) describing dynamic systems, into coherent sequences of events without the need for labeled training data. Our approach incorporates domain knowledge in the form of preconditions and effects of events, and we show that it outperforms state-of-the-art supervised learning systems on the task of reconstructing RoboCup soccer games from their commentaries.

## 1   Introduction

Natural language understanding requires the ability to translate individual sentences into representations of the underlying entities, properties, relations, and actions. It further requires the ability to combine the representations of individual sentences into a coherent whole. For example, determining from a commentary of a soccer game who has possession of the ball at any given moment requires one to infer numerous events that may not have been explicitly mentioned such as unsuccessful passing, kicking the ball, or stealing the ball.

This paper argues that the coherence of natural language text provides a strong bias that can be exploited when learning to understand language. One particularly simple form of coherence requires that events can only take place when their preconditions are met. For example, a soccer player cannot kick the ball unless he is currently in possession of the ball. We present an approach that incorporates such domain knowledge into the process of learning to understand sports commentaries without annotated data.

We use human commentaries of four championship games in the RoboCup simulation league [6] and map each narrative to a sequence of events such as *pass, kick, steal, offside* along with their arguments. Soccer commentaries differ from more complex narratives in that they report a linear sequence of events that unfolds over time. In a soccer commentary, each sentence results in an incremental update of the overall representation, or discourse model [24, 11, 9], making it possible to reconstruct the original temporal sequence of events.

In this paper we show how a small amount of domain knowledge and a bias towards coherent discourse models can be used to learn how to interpret narratives. The domain knowledge we exploit encodes the preconditions and effects of event types using a STRIPS-like framework [8]. The logical representation of knowledge allows structured modeling of event types rather than modeling propositional state transitions. In contrast to other recently proposed approaches (e.g., [25, 6, 12]), we require neither labeled training data nor an agent that receives indirect supervision by interacting with a physical environment (e.g., [3, 23]). These are often very hard, if not infeasible, to create.

Our approach outperforms the state-of-the-art approach of [6], which uses annotated data on understanding RoboCup soccer commentaries, even when extended to incorporate similar domain knowledge at inference time.

Similar to [6], we formulate language understanding as a classification problem in which we have to predict events from individual sentences. Given a sentence, our classifier assigns a score to each possible interpretation.[1] We interpret these scores as utilities and use dynamic programming to find the sequence of events that has maximal utility. In contrast to other approaches, our classifier also receives as input our guess of the current state of the world. This allows it to penalize events that violate domain constraints (because their preconditions are not met in what we assume to be the current state of the world). Our classifier is trained in an iterative fashion that is reminiscent of self-training or hard EM [2]. Starting from an initial guess of correct events for sentences and current states, we iteratively retrain according to the current version of the classifier. At test time, we use a dynamic programming approach to find a coherent sequence of events that has maximal utility according to the learned scores of events.

---

[1]An alternative approach (to be explored in future work) might treat this task as a complex sequence labeling problem where each element of the sequence corresponds to an individual sentence and the label corresponds to the predicted event.

### 1.1 Related Work

There are other approaches that learn to map natural language text to meaning representations. Some (e.g., [25]) use annotated labeled data and are focused on texts describing facts rather than describing the dynamics of a system. Most similar to our approach are [3, 6], which map narratives to a sequence of events with applications in understanding instructions or generating commentaries. [3] use reinforcement learning and have access to a physical environment to provide supervision for assigning rewards to selecting events. [6] have access to the actual events of soccer games and use a mapping between commentaries and real events of the game for training. In contrast, our approach uses prior knowledge about events, does not have access to the actual events that occurred in the soccer game, does not interact with a real physical environment, and uses binary classifiers instead of reinforcement learning.

Several approaches introduce reasoning, learning, or planning algorithms in a probabilistic logical framework to model events. Planning and reasoning approaches, unlike our approach, assume that the probability distributions over events are known. Exact reasoning approaches (e.g., [1, 18]) are not feasible for our problem, because they usually consider all possible event sequences. Approximate reasoning approaches (e.g., sampling event sequences of narratives [10]) are still too expensive. Probabilistic planning approaches [22] usually find the most likely plan given initial and goal states. Learning approaches [7, 26] also compute probability distributions over different events, but, unlike us, use labeled training data, and train their classifier to predict single probabilistic events independently rather than accumulating information from selected events using an inference subroutine.

Our approach is also related to research in planning and, to a greater degree, activity recognition. Some approaches (e.g., [13]) use logical elements and symbolic reasoning, but are not able to rank multiple plans that are consistent with the text. Other approaches use probabilistic reasoning (e.g., Bayesian networks in [5] or Hidden Markov Models (HMMs) in [4]). Other approaches (e.g., [19, 14]) incorporate learning and reasoning in dynamic models such as HMMs and Markov decision processes. These approaches are usually augmented with annotated labeled data and do not use logics to model the domain. Most recently, [20] recognize activities by applying relational inference and learning (with noisy GPS information as training labels). They neither improve initial estimates of labels nor use consistency checking. Moreover, using hard and soft constraints, they augment their system with richer (more expensive) prior knowledge compared to our few event descriptions.

## 2 Problem Definition

The problem that we address here is to find the best sequence of interpretations for the sentences of a natural language narrative. In our approach, we additionally have access to domain knowledge about events in terms of their preconditions and effects.

### 2.1 Natural Language Narratives

A narrative is a text that describes the (linear) temporal evolution of a system as a sequence of sentences in natural language. Examples of such narratives are commentaries, stories, reports, and instructions. In a narrative, each sentence leads to an incremental update of the overall representation, or discourse model. Specifically, a narrative is a sequence of length $T$ of sentences $\langle w_1, w_2, \ldots, w_T \rangle$. Throughout this paper we work with English commentaries of final games in the RoboCup soccer simulation league, taken from [6].

**Sentences:** Every sentence in the narrative is either an observation about the state of the system (e.g., "Offside has been called on the Pink team.") or a change that happens in the state of the system (e.g., "Pink9 tries to kick to Pink10, but was defended by Purple3."). Because the meaning representation language defined by [6] provides a number of event types, the above sentence could be interpreted as *passing* between players, *unsuccessful passing* between players, a player *kicking* the ball, or a player *defending* the other player.

**State of the world:** The state of the world (i.e., the soccer game) changes over time. For example, after the sentence "The pink goalie kicks off to Pink2.", a (partial) description of the current state of the game would include the fact that 'Pink2 has possession of the ball'.

### 2.2 Meaning Representations

Similar to [6], we use a logical language to model meaning representations in a structured way. Our meaning representation language consists of a finite set of state predicates, events, and their arguments corresponding to the elements of the domain. Although our domain is finite, this structured representation allows us to capture general domain knowledge about event types.

**Representation Language:** We use a logical language to represent state predicates, events, and entities. The language consists of a finite set of constants (players $Pink1$, $Purple1$, $Pink2$, $Purple2$), variables (e.g., $player_1, player_2$), state predicates (e.g., *holding(player,ball)*, *atCorner()*, *atPenalty()*), and event types (e.g., $pass(player_1, player_2)$, $kick(player1)$, $steal(player1)$).

**Definition 1.** The language $\mathcal{L}$ of our meaning representation framework is a tuple $\mathcal{L} = (C, V, F, E)$ consisting of

- $C$, a finite set of domain entities,
- $V$, a finite set of variables,
- $F$, a finite set of state predicates, and
- $E$, a finite set of event types.

Event types (e.g., $kick(player_1)$) and state predicates (e.g., *holding(player,ball)*) are syntactically represented with a name together with a list of arguments. For event types and state predicates, these arguments are all variables rather than constants. A ground event is an instantiation of an event type, meaning that the variables are replaced with constants. Similarly, a ground predicate is an instantiation of a state predicate. For example, $pass(player_1, player_2)$ is an event type because $player_1$ and $player_2$ are variables, whereas $pass(Pink1, Pink2)$ is a ground event because $Pink1$ and $Pink2$ are constants.

A (complete) state $s$ in this framework is defined as a full assignment of $\{true, false\}$ to all possible groundings of the predicates in $F$. However, at any particular time step, it is generally the case that the values of many state predicates are unknown. A belief state $\sigma$ is a set of (complete) states that hold in a particular time step. It can also be interpreted as a conjunction of those ground predicates whose truth values are known.

**Event Semantics:** Prior knowledge is provided in the form of a small number of event type descriptions in a logical format. Semantically, an event type either describes a belief state (e.g., $corner()$) or deterministically maps a belief state to a new belief state (e.g., $pass(player_1, player_2)$). The semantics of an event type is described with STRIPS preconditions and effects. These semantic descriptions are in fact the prior knowledge provided to our system.

We use our logical language to describe *event types* rather than *ground events*. Describing event types stands in contrast to describing all possible state-to-state transitions. Hereinafter, for simplicity, we use the terms 'event' instead of 'ground event', and 'state' instead of 'belief state'.

**Definition 2.** Let $e(\vec{x})$ be an event type and $\vec{x}$ its arguments. The effect axiom for the event type $e(\vec{x})$ is represented as $\langle e(\vec{x}), \textit{Precond}(\vec{x}), \textit{Effect}(\vec{x}) \rangle$ where $\textit{Precond}(\vec{x})$ and $\textit{Effect}(\vec{x})$ are conjunctions of state predicates or their negations.

For example, the event axiom $\langle pass(player_1, player_2), holding(player_1), holding(player_2) \rangle$ describes the event type $pass$ that changes the possession of the ball from $player_1$ to $player_2$, and the event axiom $\langle kick(player_1), holding(player_1), \neg holding(player_1) \rangle$ represents that $player_1$ is no longer holding the ball after he shoots. The *holding* predicate is exclusive: if one person holds the ball,

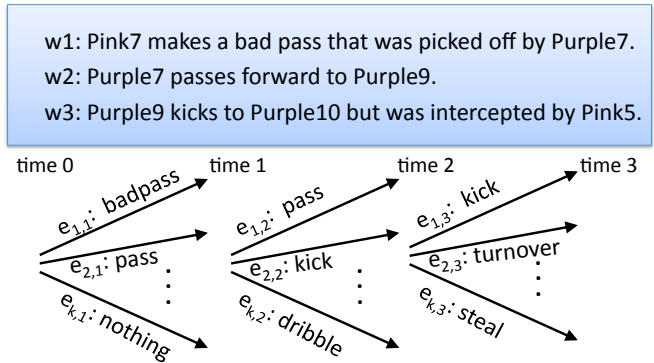

Figure 1: An example narrative as a sequence of sentences and their interpretations in the form of events.

no other person can hold it at the same time.

Here we use the frame assumption that the truth value of a gound predicate stays the same unless it is changed by an event.

Most events associated with RoboCup commentaries involve actions with the ball such as kicking and passing. There are some other events that provide game information such as whether the current state is penalty, offside, or corner. For example the event $corner$ is described as $\langle corner(), \top, \textit{atCorner}() \rangle$.

The set of event types includes a noise event called *Nothing* that has no preconditions and no effects, and hence does not alter the state of the world. The reason for including this noise event is that some narratives include sentences that are not mapped to any actual events. For example, the sentence "Today we have a nice match between the pink and the purple team." does not map to any of the described events of the soccer game. In addition, the noise event helps to fix inconsistencies that exist in the narrative. For example, a soccer commentary may be inconsistent if the commentator has missed commenting on some of the events of the game. Mapping sentences to the noise event allows some flexibility for mapping other sentences correctly.

### 2.3 Sentence Interpretations

Different events can be inferred from a natural language sentence. We assign a score to all possible events associated with a sentence. This score also depends on the current state of the world. The score corresponding to the interpretation of sentence $w$ as event $e_i$ in state $s$ represented as $P(e_i|w, s)$ for all the events $e_i$. Figure 1 shows part of a RoboCup narrative and a possible ranking of events according to their scores.

To map a narrative to a sequence of events, we first need to compute $P(e_i|w, s)$ for all ground events $e_i$ and every sen-tence $w$ in the narrative. We model this score as a logistic function [2] and learn it in an iterative learning procedure (Section 3.1). Each iteration involves estimating the label for the training examples and modifying the model accordingly.

Following [6], we assume that each sentence can be mapped to at most one event. For instance, for the sentence "Pink6 tried to pass to Pink10 but was intercepted by Purple3." the goal is to map the sentence to a single event $badPass(Pink6, Purple3)$ rather than a sequence of events like $kick(Pink6)$, then $pass(Pink6, Pink10)$, and then $badPass(Pink6, Purple3)$.

## 3 Mapping Narratives to Event Sequences

Our approach, *ITerative Event Mapping* (ITEM) (Figure 2) is built upon two subroutines of inference and iterative learning. ITEM takes as input a set of training narratives $Tr$, a test narrative $Ts$, and domain knowledge encoded in terms of the preconditions and effects of event types, $EA$ (event type axioms), in our logical language. For training, ITEM uses an iterative learning subroutine, *IterTrain* (Section 3.1, Algorithm 2), to learn the model parameter vector $\vec{\Theta}$ from the training narratives. The scores of events associated with sentences are computed according to the learned model. At test time, ITEM uses an inference subroutine (Section 3.2, Algorithm 3) to find the best event sequence for the test narrative using the learned model and the computed scores. The domain knowledge is used when initializing labels, generating training examples, and in the inference subroutine.

### 3.1 Training: Iterative Learning

Our iterative learning subroutine, *IterTrain*, is illustrated in Figure 3. Inputs to this subroutine are a set of training narratives $Tr$ and domain knowledge expressed as event type effect axioms $EA$ in our logical language $\mathcal{L}$. In our experiments (following [6]) we use three RoboCup soccer games as training data and the fourth as the test narrative. The output of *IterTrain* is a model parameter vector $\vec{\Theta}$ estimated on the training narratives. *IterTrain* trains a binary classifier to separate the correct event from the other events for every sentence and state. To model this problem as a binary classification task, we build training examples as triplets (state $s$, sentence $w$, event $e_i$) from the training narratives and assign boolean labels to them. A positive label for a training example shows that event $e_i$ is the correct interpretation of the sentence $w$ in the state $s$.

*IterTrain* consists of different modules. The *Example Generator* module generates training examples $(s, w, e)$ from the training narratives. Afterwards, *Feature Extractor* generates features $\vec{\Phi}(s, w, e)$ for each training example. Because the correct labels of training examples are not known, *Initial Label Generator* estimates initial training labels using prior knowledge. Then, in the first iteration the *Classifier* module learns the model parameters $\vec{\Theta}_1$ for the current training examples and the estimated labels. It uses logistic regression [2] to compute the binomial probability $P_{\vec{\Theta}}(e|w, s)$ that event $e$ is the correct interpretation of the sentence $w$ in state $s$.

**Algorithm 1.** ITEM(*Tr, Ts, E, EA*)
Input: training narrative *Tr*, test narrative *Ts*, event types *E*, effect axioms *EA*
1. $\vec{\Theta} \leftarrow$ *IterTrain*(*Tr, EA*)
2. *Inference*(*Ts*, $\vec{\Theta}$, *E, EA*)

**Algorithm 2.** *IterTrain*(*Tr, EA*)
Input: training narrative *Tr*, effect axioms *EA*
1. $k \leftarrow 0$, randomly initialize $\vec{\Theta}_0, \vec{\Theta}_1$
2. Repeat until $||\vec{\Theta}_{k+1} - \vec{\Theta}_k|| < \epsilon$
   (a) $k \leftarrow k + 1$
   (b) $(s_i, w_i, e_i)_{i:1..S} \leftarrow$ *ExampleGenerator*(*Tr, EA*)
   (c) **for** $i : 1$ to $S$
       i. $\vec{\Phi}_i \leftarrow$ *FeatureExtractor*($s_i, w_i, e_i$)
       ii. if $k = 1$:
           A. $l_i \leftarrow$ *InitialLabelGen*($s_i, w_i, e_i, EA$)
       iii. else: $l_i \leftarrow$ *UpdateLabels*($s_i, w_i, e_i, EA$)
   (d) $\vec{\Theta}_k \leftarrow$ *Classifier*($\vec{\Phi}_{i:1..S}, l_{i:1:S}$)
3. return $\vec{\Theta} \leftarrow \vec{\Theta}_k$

**Algorithm 3.** *Inference*(*Ts*, $\vec{\Theta}$, *E, EA*)
Input: test narrative $Ts = \langle w_1 \ldots w_T \rangle$, $\vec{\Theta}$, *E,EA*
Output: sequence of events $\langle e_1, \ldots, e_T \rangle$
1. if $t = 1$ initialize $V_{1,e_i}, S_{1,e_i}, Seq_{1,e_i}$ from Eq. 2
2. for $t = 2 \ldots T$
   (a) for $e_i$ in $E$:
       i. $e^* \leftarrow \arg\max_{e_j \in E} P_{\vec{\Theta}}(e_i|w_t, S_{t-1,e_j}) + V_{t-1,e_j} + r_{S_{t-1,e_j},e_i}$
       ii. $V_{t,e_i} \leftarrow P_{\vec{\Theta}}(e_i|w_t, S_{t-1,e^*}) + V_{t-1,e^*} + r_{S_{t-1,e^*},e_i}$
       iii. $S_{t,e_i} \leftarrow$ *Progress*($S_{t-1,e^*}, e_i$)
       iv. $Seq_{t,e_i} \leftarrow Seq_{t,e^*} + [e_i]$
3. $e_T \leftarrow \arg\max_{e_i \in E}(V_{T,e_i}), e_{1..T-1} \leftarrow Seq_{e_t,t}$

Figure 2: The ITerative Event Mapping (ITEM) algorithm finds the best event sequence corresponding to a narrative. The subroutine *IterTrain* uses the training narrative to estimate the model parameters $\vec{\Theta}$, and *Inference* finds the best event sequence for the test narrative given the learned model parameters and the event type descriptions.

At each iteration $k > 1$, the algorithm uses the learned model parameters $\vec{\Theta}_{k-1}$ and updates the previously estimated labels for the training examples using the *UpdateLabels* module. Iterations will continue until convergence, i.e., until ($||\vec{\Theta}_{k+1} - \vec{\Theta}_k|| < \epsilon$).

For testing, we use the final learned model $\vec{\Theta}$ to compute scores $P_{\vec{\Theta}}(e_i|w, s)$ of all events $e_i$ associated with every sentence $w$ and the current state $s$ in the test narrative. In the following we describe the different subroutines used for training in more detail.

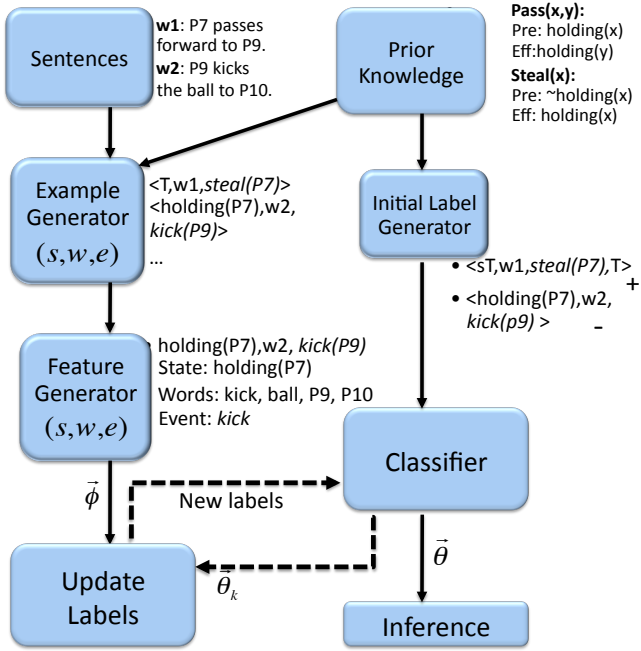

Figure 3: The architecture of our iterative learning approach *IterTrain* (Algorithm 2) to compute the model parameters $\vec{\Theta}$ and return the score of every event given the sentence and the current state.

**Training Example Generator:** The inputs to this module are training narratives consisting of a sequence of sentences $\langle w_1 \ldots w_T \rangle$ and event effect axioms. It generates training examples of the form (state $s_{i,t}$, sentence $w_t$, event $e_{i,t}$). This module first samples sequences of event types for the narrative, grounds the event types, and generates (sentence, event) pairs. It then updates the state of the world for each pair. This provides the final triplets of training examples.

To build pairs of $(w_t, e_{i,t})$ for each sentence $w_t$, we sample $N$ (10 in our experiments) event types $e_{i,t}$ uniformly from the set of event types $E$ of $\mathcal{L}$. For example, we may sample the event types *pass*, *steal*, and *kick* for the sentence "P7 passes the ball to P9"(P stands for Pink) in Figure 3. We then deterministically ground event types by assigning players as arguments to the event types. For that, we assume that we know the list of players for the soccer game. We select arguments (players in the soccer domain) for the event types from those appearing in the sentence or nearby sentences. For example, arguments for the above sentence are $P7$ and $P9$. To ground the event type, we replace the variables in the event type $pass(player_1, player_2)$ with the constants $P7$ and $P9$ (see section 4 for details). For the above sentence the ground event is $pass(P7, P9)$, where $player_1$ is replaced with $P7$ and $player_2$ is replaced with $P9$.

We then compute the state of the narrative given the sequence of ground events $evs_i = \langle e_{i,1}, \ldots e_{i,T} \rangle$ corresponding to sentences $w_1 \ldots w_T$ of the narrative. We build a new sequence of the form $\langle s_{i,0}, e_{i,1}, s_{i,1}, e_{i,2}, s_{i,2}, \ldots, e_{i,T}, s_{i,T} \rangle$. Every $s_{i,t}$ is the state of the narrative at time $t$ and is computed using preconditions and effects of event axioms. Initially, any ground predicate can be *true* i.e., $s_{i,0} = \top$. Then, at each time step $t$, we apply the effects of the event $e_{i,t}$ and update the state $s_{i,t-1}$ using a *Progress* subroutine. Therefore, the triplets $(s_{i,t-1}, w_t, e_{i,t})$ are generated given all the events and the updated states.

For example, one instantiated sequence of event types sampled for the narrative in Figure 3 is $\langle steal(P7), kick(P9) \rangle$. We update the state given the event sequence and derive $\langle \top, steal(P7), holding(P7), kick(P9), \neg holding(P9) \rangle$: If the event $steal(P7)$ happens in $s_0 = \top$, the state $s_1$ is $holding(P7)$. From this sequence we extract two training examples $(\top, w1, steal(P7), \top)$ and $(holding(P7), w2, kick(P9))$.

To update the state of the system with events we use the *Progress* subroutine. $Progress(s_{t-1}, e_t)$, takes as input an event $e_t$ and the current state $s_{t-1}$ and returns the updated state $s_t$. If the preconditions of the event $e_t$ are consistent with $s_{t-1}$, then $s_t$ is updated by applying the grounded effect axioms of the event $e_t$. Otherwise, $s_t$ is assigned to $\top$ to show that any predicate state can be true.

**Feature Extractor:** This module takes an example (sentence $w$, event $e$, state $s$) as input and returns the corresponding feature vector $\vec{\Phi} = (1, \vec{\phi}_s, \vec{\phi}_w, \vec{\phi}_e)$, where 1 is a bias term. $\vec{\phi}_s$ is a binary vector representing the state $s$ where each element in the vector represents the value of the corresponding ground predicate in the state $s$. Because we have few ground predicates in the RoboCup domain, we model states as truth assignments to ground predicates rather than as state predicates. For larger domains, we can avoid grounding by modeling predicate names and their arguments separately in the feature vector.

$\vec{\phi}_w$ is a binary vector representing the sentence $w$, where each element in the vector shows the presence of the corresponding word of the vocabulary or the player names in the sentence. $\vec{\phi}_e$ is a binary vector representing the type of the event $e$, where each element in the vector represents the presence of the corresponding event type in the training example. Arguments of the event $e$ are encoded in the 'sentence' part, $\vec{\phi}_w$. Our approach learns event types and assigns the players extracted from the sentence as arguments to the event types.

**Initial Label Generator:** This module distinguishes us from learning approaches that use manually labeled data. We only use automatically generated labels. This module takes as input an example (state $s$, sentence $w$, event $e$) and provides an estimate of the actual label. The intuition behind a positive label is that event $e$ is a correct interpretation of the sentence $w$ given that the state is $s$.

In the first iteration, this module uses the event type axioms and automatically assigns labels to the training examples. More specifically, an example $(s, w, e)$ will be assigned a positive label if the preconditions of event $e$ are consistent with the current state $s$ and if the name of event $e$ has low edit distance to a word in the sentence $w$. This is only a rough estimation because for example the verb of the sentence "P7 passes the ball to P9." is *kick* where the correct event type is *pass*, which has a large edit distance to the words of the sentence. This way, there can be more than one positive example corresponding to every sentence. However, most of training examples are assigned a negative label because for most of the events in training examples the above conditions are not satisfied.

For instance, the training example $(\top, w1, steal(P7))$ has been assigned a positive label because $steal(P7)$ is feasible in $s_0 = \top$. However, the example $(holding(P7), w2, kick(P9))$ has been assigned a negative label because $kick(P9)$ is not feasible if $s=holding(P7)$.

**Update Labels:** In subsequent iterations $k > 1$, new labels are proposed by applying the current classifier to training examples. This module uses the learned model parameter vector $\vec{\Theta}_{k-1}$ from the previous iteration $k-1$ and computes the score of each training example $(s, w, e)$ using the logistic function:

$$P_{\vec{\Theta}}(e|w, s) = 1/(1 + \exp(-\vec{\Theta}_{k-1} \times \vec{\Phi}(s, w, e))) \quad (1)$$

An example $(s, w, e)$ will be assigned a positive label if its score is higher than the score of $(s, w, e')$ for every other event $e'$.

**Classifier:** At each iteration $k$, the *classifier* takes training examples $(s, w, e)$ together with the generated labels for that iteration and returns the model parameters $\vec{\Theta}_k$. This module first removes some of the negative examples randomly to balance the number of training examples so that the number of positive and negative labels is comparable. It then trains a linear classifier (*logistic regression* [2]) to compute the binomial probability score $P_{\vec{\Theta}}(e|w, s)$ for events $e$ given the sentence $w$ and state $s$. We model this score as a logistic function (Equation 1) where $\vec{\Phi}$ is the feature vector associated with training examples and $\vec{\Theta}_k$ is a vector of model parameters that we want to learn. The output of the classifier is the model parameter vector $\vec{\Theta}_k = (\theta_1, \vec{\theta}_s, \vec{\theta}_w, \vec{\theta}_e)$, which is used to compute the score of events.

The above steps continue until the model parameters converge, i.e., when $||\vec{\Theta}_k - \vec{\Theta}_{k-1}|| < \epsilon$. Our iterative learning approach takes ideas from hard EM. In analogy to the maximization step, we use our logistic regression classifier and optimize to find the model parameters. In analogy to the expectation step, we compute the new scores of training examples with the logistic function and update the labels of the training examples. Therefore, similar to EM-like approaches, our approach converges to a local minimum. We call the final learned model, which will be used for testing, $\vec{\Theta}$.

At test time, our approach selects parameters (players in our example) for the learned event type from those appearing in the sentence or nearby sentences. It grounds all the possible event types associated with the sentence using the derived parameters. It uses the final learned model $\vec{\Theta}$ to compute the score of events $e_i$ corresponding to the test sentence $w$ in the state $s$ using the logistic function. Finally, $P_{\vec{\Theta}}(e_i|w, s)$ is derived by normalizing the scores of all events $e_i$ corresponding to the sentence $s$ and the state $s$.

### 3.2 Testing: Inference with Dynamic Programming

The *Inference* subroutine (Algorithm 3) takes as input the test narrative $\langle w_1, \ldots, w_T \rangle$, the event effect axioms *EA*, and the learned model parameter vector $\vec{\Theta}$. It then returns the best event sequence corresponding to the test narrative and our model using a Viterbi-like [17] dynamic programming approach.

Intuitively, the inference subroutine selects the event that is most likely and feasible in the state of the narrative. Our dynamic programming approach maximizes the utility of selecting the event $e_i$ at time $t$. It uses following recursive relations to find the best event sequence corresponding to the narrative.

Here $V_{t,e_i}$ corresponds to the maximum utility of selecting event $e_i$ in time step $t$. This value is initialized with the score of events for the first sentence and the initial state. These scores $P_{\vec{\Theta}}(e_i|w_1, s_0)$ are derived using the learned parameter vector $\vec{\Theta}$. If the preconditions of $e_i$ are not consistent with the current state $s$, we penalize the value of choosing this event using a penalty function $r_{s,e_i}$, which is a real number between 0 and $-1$. In our experiments, we set this penalty function to $-1$ to penalize events with higher scores more than the events with lower scores. $e^*$ is the event selected at previous time step to achieve the maximum utility at time $t$ by selecting the event $e_i$. The current state $S_{t,e_i}$ is derived by *Progressing* state $S_{t-1,e^*}$ with event $e_i$.

Initialize for all $i$: \hfill (2)
$$V_{1,e_i} = P_{\vec{\Theta}}(e_i|w_1, s_0)$$
$$S_{1,e_i} = Progress(s_0, e_i)$$

Iterative step for all $t \in 2..T, e_i$:
$$e^* = \arg\max_{e_j} P_{\vec{\Theta}}(e_i|w_t, S_{t-1,e_j}) + V_{t-1,e_j} + r_{S_{t-1,e_j},e_i}$$
$$V_{t,e_i} = P_{\vec{\Theta}}(e_i|w_t, S_{t-1,e^*}) + V_{t-1,e^*} + r_{S_{t-1,e^*},e_i}$$
$$S_{t,e_i} = Progress(S_{t-1,e^*}, e_i)$$

where $s_0 = \top$ and $r_{s,e_i} = \begin{cases} -1 & Precond(e_i) \not\models s \\ 0 & \text{otherwise} \end{cases}$ is a function for penalizing infeasible events.

To update the best sequence of events, we use the following recursive formulas. $Seq_{t,e_i}$ corresponds to the best sequence of events if event $e_i$ is selected at time $t$.

$$Seq_{1,e_i} = [e_i]$$
$$Seq_{t,e_i} = Seq_{t-1,e^*} + [e_i] \qquad (3)$$

The event sequence $Seq_{t,e_i}$ is updated by keeping a pointer to the previously best selected event in the recursive step. Finally, the best sequence of events is derived by selecting the event at time $T$ that has the highest value and backtracking recursively, i.e., $e_T = \arg\max_{e_i \in E}(V_{T,e_i})$ and $e_{1..T-1} = Seq_{T,e_T}$.

Please notice that the error of finding the best sequence propagates further with time. Exact approaches for this problem (of finding the best path) are infeasible computationally given the length of the narratives. Therefore, we use our approximate Viterbi-like approach, which is more tractable. However, an exact method (considering all the possible paths) can perform better when feasible (e.g., when the narratives are very short).

## 4 Experiments

In this section we evaluate how well our algorithm, ITEM, can map narratives to a sequence of events. We work with RoboCup soccer commentaries [6]. The task is to compute the accuracy of the mapped event sequence with respect to a gold standard event sequence. We compare the accuracy of our approach with baseline algorithms and a state-of-the-art approach that uses annotated labeled data. Our experiments demonstrate that prior knowledge about event effect axioms alleviates the need for labeled data.

### 4.1 RoboCup Soccer Commentaries

We use the RoboCup soccer commentaries dataset [6]. The data is based on commentaries of four championship games of the RoboCup simulation league that took place from 2001 to 2004. Each game is associated with a sequence of human comments in English and their meaning representations. There are in total of 1872 comments where the 2001, 2002, 2003, and 2004 games have 672, 459, 398, and 343 comments, respectively.

We add a unary predicate, *holding*, and a noise event, *Nothing*, to the meaning representations of [6]. In addition, we manually describe effect axioms for event types. Event types for the soccer commentary include actions with the ball or other game information. In total there are 16 event types among which there are 3 with two arguments, 4 with one argument, and 10 with zero arguments. The number of state predicates in the domain is 10. We assign constants as player names. In total there are 24 constants.

In addition to the English comments about the game, the original dataset includes real events (represented in the same meaning representation language) that happen in the original game tagged with time. Our ITEM algorithm does not use this event log of the game. This makes us different from the approach in [6], which uses annotated labeled data in the form of a mapping between natural language comments and the real events that occurred within 5 time steps of when the comments were recorded.

### 4.2 Mapping Sentences to Events

We use gold-standard labels in the dataset as the correct event corresponding to the sentence, only for evaluation purposes. We evaluate the accuracy of the output sequence of events by computing the proportion of the events that have been correctly assigned to the sentences. We report the results for every game using the other games for training. We also report the micro-averaged accuracy over all examples. We compare the accuracy of our approach with [6] and several baselines.

#### 4.2.1 Our approach (ITEM)

Our classifier trains on three RoboCup games (e.g., 2001, 2002, and 2003) and tests on the last game (e.g., 2004). We run *IterTrain*, using $N = 10$ samples for each sentence, and compute the model parameter vector. The average number of training examples per iteration is about 700. To generate training examples, we use the *Example Generator* module to sample event types for sentences, ground event types based on the arguments of the sentence (player names), and update the states based on the events. There are cases where we miss arguments, for example when the player name is mentioned as "Pink Goalie" rather than "Pink1". There are also cases in which the sampled event type requires more arguments than the player arguments extracted for the sentence. For example, the sentence "He kicks the ball to P10" has one player argument *P10*. In these cases we use the last argument selected for the previous sentence as an argument of the event type. Also, there are cases in which the sentence has more arguments than the event type. In such cases we consider different pairs of argument players as arguments of the event types. Currently, we deterministically assign arguments, but this can be easily replaced with building a list of candidates among which our approach has to choose.

After generating examples, we extract features for each example. The feature vector $\vec{\Theta}$ has 250 elements consisting of 16 event types, 195 words, and 38 ground state predicates. We use *Initial Label Generator* to generate initial labels using event effect axioms and computing edit distance of the event name and words in the sentence. A training example

$(s, w, e)$ gets a positive label if $e$ is feasible in state $s$ and its name has low edit distance ($\leq 3$) to at least a word in the sentence. The initial labels are later changed in subsequent iterations using the learned classifier. Using the learned model parameter vector, we compute the score of every event associated with a sentence in the test narrative. Finally, we apply our inference subroutine and find the best event sequence for the test narrative.

We first show in Graph 4 that our iterative learning approach converges and improves the accuracy of labels generated for the training examples. We report the average accuracy of labels at each iteration in terms of an $F_1$ measure over all four training scenarios.

Table 2 shows some examples of our correctly and incorrectly predicted labels. Our approach can identify the interpretation of the sentence "Purple10 kicks to Purple11" by mapping it correctly to $pass$ rather than $kick$, showing that the final model does not disambiguate based on the similarity of the event name and the verb name despite the initialization. Our incorrectly predicted labels show that distinguishing between $turnover$ and $badPass$ event types is hard: $badPass$ refers to the event that the agent is trying to pass but the pass mistakenly goes to the next team's player, whereas $turnover$ refers to the event that the player accidentally loses the ball. In addition, Table 1 shows the accuracy of the mapping derived by our approach.

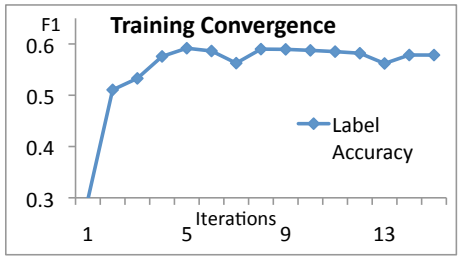

Figure 4: Convergence of the accuracy of estimated labels vs. number of iterations of training according to $F_1$ measure.

### 4.2.2 Comparison to Baselines

We compare our approach with several different baselines. For each sentence, these baselines select events with specific properties that can be candidate interpretations of the sentences.

**Uniform:** The first baseline, *Baseline-0* selects an event per sentence from a uniform distribution over events. This baseline shows how difficult the RoboCup soccer commentary is, and as one can see from Table 1 this random selection performs poorly on the dataset.

**Uniform+heuristics:** The second group of baselines show how heuristics for event selection help. *Baseline-1a* selects uniformly among event types whose names have a small edit distance ($\leq 3$) to at least one word in the sentence. It then assigns players from the sentences to the arguments of the event types. We call these events *similar* events. *Baseline-1b* selects uniformly among event types whose arity is equal to the number of arguments in the sentence. We call these events *same-arity* events. Table 1 shows the results of these baselines. If no event type has these properties, we randomly select among all the possible event types. Although the accuracy of these baselines is significantly lower than our ITEM algorithm, they show a significant improvement over the random selection of *Baseline-0*.

| Approach | 2001 | 2002 | 2003 | 2004 | Avg. |
|---|---|---|---|---|---|
| uniform | | | | | |
| Baseline-0 | .062 | .062 | .062 | .062 | .062 |
| uniform+heuristics | | | | | |
| Baseline-1a | .688 | .464 | .628 | .437 | .574 |
| Baseline-1b | .625 | .570 | .718 | .720 | .648 |
| prior knowledge+heuristics | | | | | |
| Baseline-2a | .693 | .455 | .640 | .454 | .579 |
| Baseline-2b | .629 | .575 | .826 | .737 | .677 |
| Baseline-3 | .687 | .474 | .658 | .478 | .590 |
| prior knowledge (no labeled data) + our method | | | | | |
| **Our ITEM** | **.799** | .681 | **.867** | .769 | **.779** |
| annotated labeled data + WGIM | | | | | |
| WGIM[6] | .721 | .664 | .683 | .746 | .703 |
| prior knowledge+ annotated labeled data+WGIM | | | | | |
| WGIM[6]-*Inference* | .767 | **.721** | .638 | **.798** | .734 |

Table 1: The micro-average accuracy of different approaches for mapping RoboCup narratives to event sequences. Our approach ITEM with no annotated data shows higher accuracy compared to other algorithms. The results suggest that knowledge about event axioms together with iterative learning replaces the need for annotated labeled data and that prior knowledge improves learning with annotated data.

**Prior knowledge + Heuristics:** The next baseline examines the effect of prior knowledge without learning. *Baseline-2a* uses event effect axioms over *similar* events. This baseline applies the *inference* subroutine (Algorithm 3) with a uniform distribution over *similar* events. *Baseline-2b* applies the *inference* subroutine, but with a uniform distribution over *same-arity* events to the sentence. We apply prior knowledge during the *inference* subroutine over two previous heuristics. Our results suggest that prior knowledge helps, but the way in which prior knowledge is incorporated is important. Applying prior knowledge over uniform selection of arity events improves the accuracy up to 3%, but adding prior knowledge over uniform selection of *similar* events only improves the accuracy by 0.3%. Notice that our ITEM approach still has significantly higher accuracy compared to these baselines. This is because it uses prior knowledge together with iterative learning.

The last baseline, *Baseline-3* uses prior knowledge together with both heuristics. It selects *similar* and *same-arity*

| Correctly Predicted Events | |
|---|---|
| Sentence | Predicted event |
| Purple10 makes a quick pass to Purple11 on the side. | $pass(purple10, purple11)$ |
| Purple11 tries to pass back but was picked off by Pink5. | $badPass(purple11, pink5)$ |
| Pink5 made a bad pass that was intercepted by Purple10. | $badPass(pink5, purple10)$ |
| Purple10 kicks to Purple11. | $pass(purple10, purple11)$ |
| Purple11 threads a nice pass to Purple10 near the penalty area. | $pass(purple11, purple10)$ |

| Incorrectly Predicted Events | | |
|---|---|---|
| Sentence | Correct event | Predicted event |
| Pink6 steals the ball from Purple6. | $steal(pink6)$ | $badPass(pink6, purple6)$ |
| Pink6 tries to dribble toward the goal but turns the ball over to Purple3. | $turnover(pink6, purple3)$ | $badPass(pink6, purple3)$ |
| Purple6's pass was defended by Pink6. | $defense(purple6, pink6)$ | $Pass(purple6, pink6)$ |

Table 2: (*top*) Some sentences in the 2001 game and our responses that predicted the correct event. (*bottom*) Some examples in the dataset that were incorrectly predicted by our approach.

events for a sentence. If no event type is selected it considers all the possible events. It then applies the *inference* subroutine with a uniform distribution over selected events. Surprisingly, the results of table 1 show that the accuracy of combining both heuristics together with prior knowledge is lower than the accuracy of arity selection with prior knowledge. This shows that a naive way of incorporating prior knowledge together with heuristics does not help.

**Annotated Labeled data:** We also compare our algorithm with the state-of-the-art approach, WGIM [6]. These results show that our iterative learning method alleviates the need for labeled data collection.

In the next experiment, we augment the WGIM approach with the same event axioms (WGIM-*Inference*). We apply *inference* where the transition model for every sentence is derived from the WGIM approach. At each step, we select the event that has the highest score according to WGIM and is feasible in the current state. The current state is updated according to the prior selected events in the sequence. The accuracy of WGIM-*Inference* is still lower than our ITEM approach. This is because the original WGIM does not learn the event scores given the current state. However, our results show that adding prior knowledge to WGIM improves WGIM's accuracy.

## 5 Discussion and Future Work

In this paper we have introduced an approach for mapping RoboCup soccer narratives to sequences of events without the need for any annotated data. We show that knowledge about event models together with a careful design of representation, inference, and iterative learning alleviates the need for annotated data. We show that this iterative learning approach achieves superior accuracy compared to both heuristics and a state-of-the-art approach that uses labeled data. The transferable insight of the current paper is that a small number of partial descriptions of events enables improved understanding with little or no labeled data (still assuming a body of examples, only possibly unlabeled). Our approach takes advantage of a logical structure for event descriptions that is precise and easy to use because we describe a small number of event types rather than all the possible transitions in the system. In addition, our approach takes advantage of an iterative learning algorithm to build a precise, strong, easy to use, and easy to engineer model.

We show that by collecting prior knowledge about events we do not need to annotate every sentence in the domain. It is usually very hard to scale labeled data because we need to generate more labels to be able to understand larger texts. However, if we collect prior knowledge for a specific context, we can understand large texts in that context. We plan to extend our approach to understanding other narratives and answering questions about them. Specifically, we would like to work with stories in the Remedia corpus or Weblog stories [16]. In this setting, we plan to use VerbNet [21], a comprehensive verb lexicon that contains semantic information such as preconditions, effects, and arguments of about 5000 verbs. Our current approach uses a simple bag-of-words model to represent sentences. We do not use tools from natural language processing literature to process sentences. The reason is that our domain contains simple sentences that does not require more complex text processing. In the general setting, we plan to extend our simple bag-of-words model with syntactic parsing of the sentences.

One shortcoming of our approach is that we assume the input narrative corresponds directly to a linear sequence of events. This may not be the case in genres where rhetorical relations such as *explanation* or *elaboration* occur [15]. Even in our domain, it is possible that some events have been missed or the sentences are represented in partial order. We intend to explore the use of probabilistic and partial order planning approaches to infer the event sequences. We also plan to investigate multi-class classification approaches instead of binary classification, so that the goal is to learn a multinomial probability distribution over different events corresponding to a sentence and the state, and to cast the problem as sequence labeling problem where each element of the sequence corresponds to an individual sentence and the label to the predicted event.

**Acknowledgements**

We thank Dan Roth and Leora Morgenstern for their insightful comments on this paper. We are also grateful to David Chen and Raymond Mooney for sharing with us the RoboCup dataset and their algorithm's results. In addition, we would like to thank the anonymous reviewers for their helpful comments on improving the current paper. This work was supported by NSF IIS grant 09-17123 titled "Scaling Up Inference in Dynamic Systems with Logical Structure", and NSF IIS grant 09-68552 titled "SoCS: Analyzing Partially Observable Computer-Adolescent Networks". Julia Hockenmaier is supported by NSF CAREER award 1053856 (Bayesian Models for Lexicalized Grammars) and NSF IIS INT2-Medium (0803603):